\PassOptionsToPackage{pdftex,dvipsnames}{xcolor}
\documentclass[letterpaper, 10 pt, conference]{ieeeconf}  

\IEEEoverridecommandlockouts                             
\usepackage{graphics}
\usepackage{epsfig} 
\usepackage{mathptmx} 
\usepackage{times} 
\usepackage{amsmath} 
\usepackage{amssymb}  
\usepackage{subcaption}
\usepackage{graphicx}
\usepackage{float}
\usepackage{bm}
\usepackage{caption}
\usepackage{mathtools}
\usepackage{stfloats}
\usepackage{mathrsfs}
\usepackage{multirow}
\usepackage{makecell}
\usepackage{vcell}
\usepackage{booktabs}
\usepackage{diagbox}
\usepackage{csquotes}
\usepackage{tikz}
\usepackage{lipsum}
\usepackage{schemata}
\usepackage{colortbl}
\usepackage[mathscr]{euscript}

\DeclareMathOperator*{\argmax}{arg\,max}

\usepackage{algorithm}
\usepackage[noend]{algpseudocode}

\algrenewcommand\algorithmicforall{\textbf{for each}}
\algrenewcommand\algorithmicindent{.8em}

\algdef{SE}[DOWHILE]{Do}{doWhile}{\algorithmicdo}[1]{\algorithmicwhile\ #1}%
\algnewcommand\algorithmicforeach{\textbf{for each}}
\algdef{S}[FOR]{ForEach}[1]{\algorithmicforeach\ #1\ \algorithmicdo}

\usetikzlibrary{decorations.pathreplacing,calc}
\newcommand{\tikzmark}[1]{\tikz[overlay,remember picture] \node (#1) {};}

\newcommand*{\AddNote}[4]{%
    \begin{tikzpicture}[overlay, remember picture]
        \draw [decoration={brace,amplitude=0.5em},decorate,ultra thick,black]
            ($(#3)!(#1.north)!($(#3)-(0,3)$)$) --  
            ($(#3)!(#2.south)!($(#3)-(0,3)$)$)
                node [align=center, text width=2.5cm,  anchor=west,  rotate=90, pos=0.5, anchor=north, shift={(0,-0.3)}] {#4};
    \end{tikzpicture}
}%

\usepackage[font=footnotesize]{caption}

\usepackage{hyperref}
\hypersetup{
colorlinks=true,
linkcolor=blue,
filecolor=magenta,
urlcolor=blue,
}
\usepackage{cite}

\title{\LARGE \bf
ARTEMIS: AI-driven Robotic Triage Labeling and Emergency Medical Information System
}
\author{Revanth Krishna Senthilkumaran$^{1}$, Mridu Prashanth$^{2}$,  Hrishikesh Viswanath$^{2}$, \\Sathvika Kotha$^{2}$,  Kshitij Tiwari$^{2}$, and Aniket Bera$^{2}$
\\ \textit{Purdue University, USA}
\thanks{$^{1}$ School of Electrical and Computer Engineering, Purdue University, West Lafayette, IN 47907, USA
{\tt\small senthilr@purdue.edu}}%
\thanks{$^{2}$ Department of Computer Science, Purdue University, West Lafayette
{\tt\small \{mprasha,hviswan,kotha8,aniketbera\}@purdue.edu}}%
}

\begin{document}
\maketitle
\thispagestyle{empty}
\pagestyle{empty}



\begin{abstract}
Mass casualty incidents (MCIs) pose a significant challenge to emergency medical services by overwhelming available resources and personnel. Effective victim assessment is key to minimizing casualties during such a crisis. We introduce ARTEMIS, an AI-driven Robotic Triage labeling and Emergency Medical Information System, to aid first responders in MCI events. It leverages speech processing, natural language processing, and deep learning to help with acuity labeling. This is deployed on a quadruped that performs victim localization and preliminary injury severity assessment. First responders access victim information through a Graphical User Interface (GUI) that is updated in real-time. For validation, an algorithmic triage protocol is proposed, using the Unitree Go1 quadruped. The robot identifies humans, interacts with them, gets vitals and information, and assigns an acuity label. Simulations of an MCI in software and a controlled environment outdoors were conducted. The system achieved a triage-level classification accuracy of over 74\% on average and 99\% for the most critical victims, i.e. level 1 acuity, outperforming state-of-the-art deep learning-based triage classification systems. In this paper, we showcase the potential of human-robot interaction in assisting medical personnel in MCI events.
\end{abstract}

\section{Introduction}

Mass Casualty Incidents (MCIs) globally cause significant loss of life and infrastructure damage. These incidents, contributing to tens of thousands of casualties worldwide, highlight the urgency for effective management. In 2023, the United States experienced nine MCIs related to climate crisis, leading to over \textdollar{}1 billion in damages and 99 deaths. The frequency of MCIs has been rising over recent decades \cite{santos2018ascertainment, kishore2018mortality}, posing challenges for first responders, who must manage overwhelming caseloads with limited resources under time constraints. Efficient triage systems are essential to prioritize treatment and manage resources effectively in such high-pressure situations and robotics.


\begin{figure}[htb]
\centering
 \begin{subfigure}[b]{1\linewidth}
     \centering
    \includegraphics[width=\linewidth]{title1.pdf}
     \label{fig:title1}
 \end{subfigure}
 \hfill
 \begin{subfigure}[b]{0.24\linewidth}
     \centering
     \includegraphics[width=\linewidth]{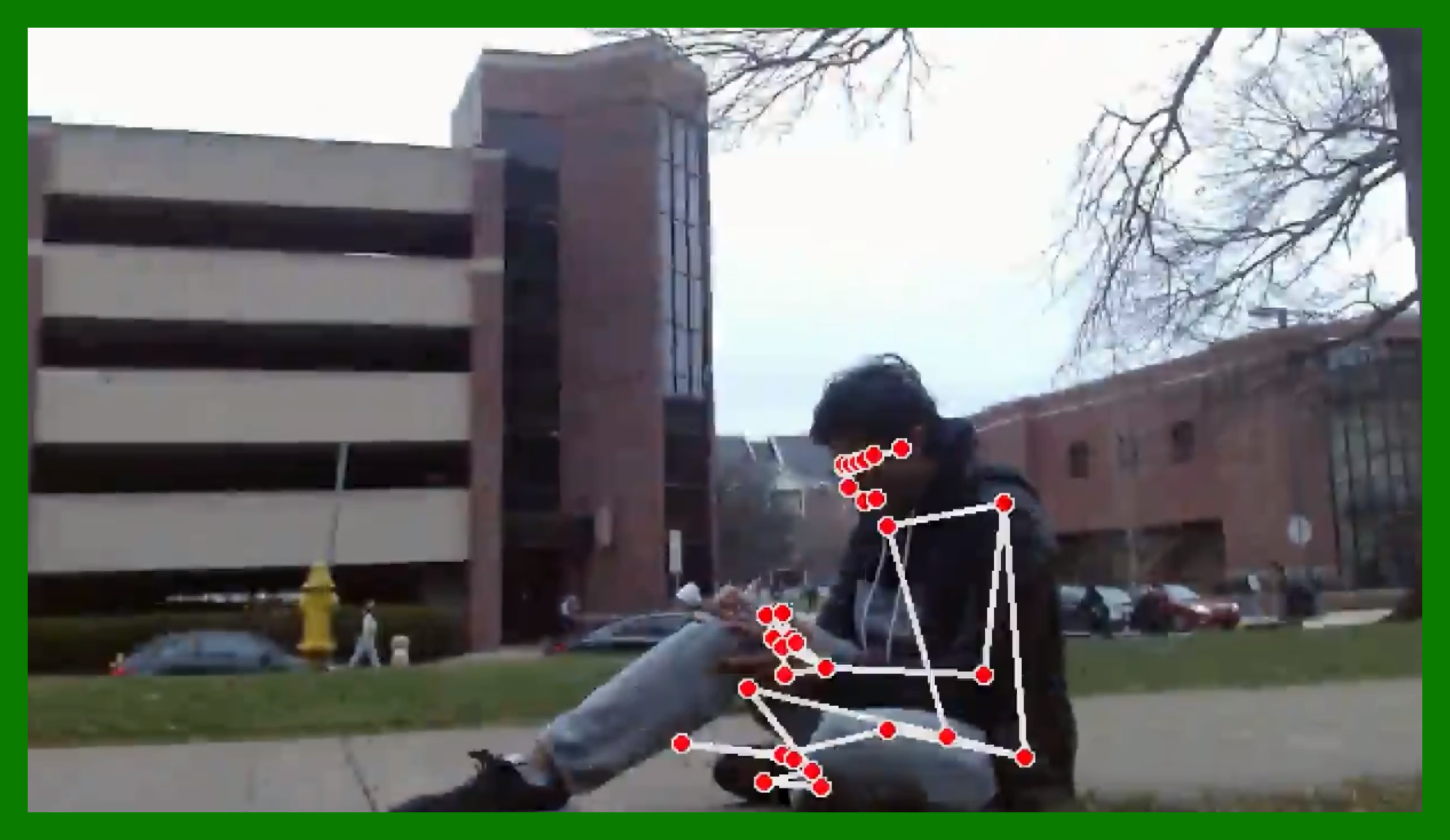}
     \caption{Far away pose}
     \label{fig:far1}
 \end{subfigure}
 \hfill
 \begin{subfigure}[b]{0.24\linewidth}
     \centering
     \includegraphics[width=\linewidth]{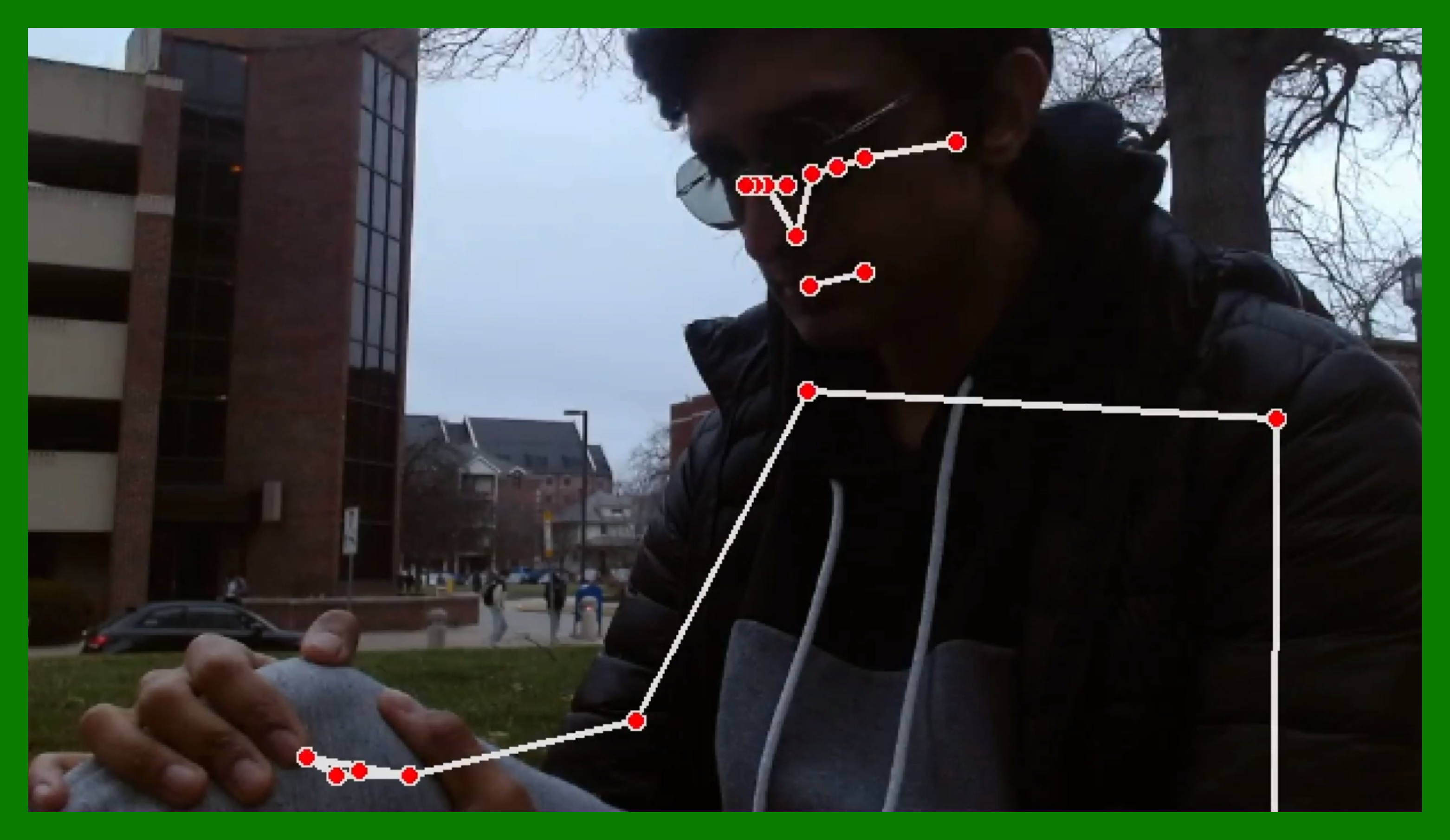}
     \caption{Close-up pose}
     \label{fig:close1}
 \end{subfigure}
  \hfill
 \begin{subfigure}[b]{0.24\linewidth}
     \centering
     \includegraphics[width=\linewidth]{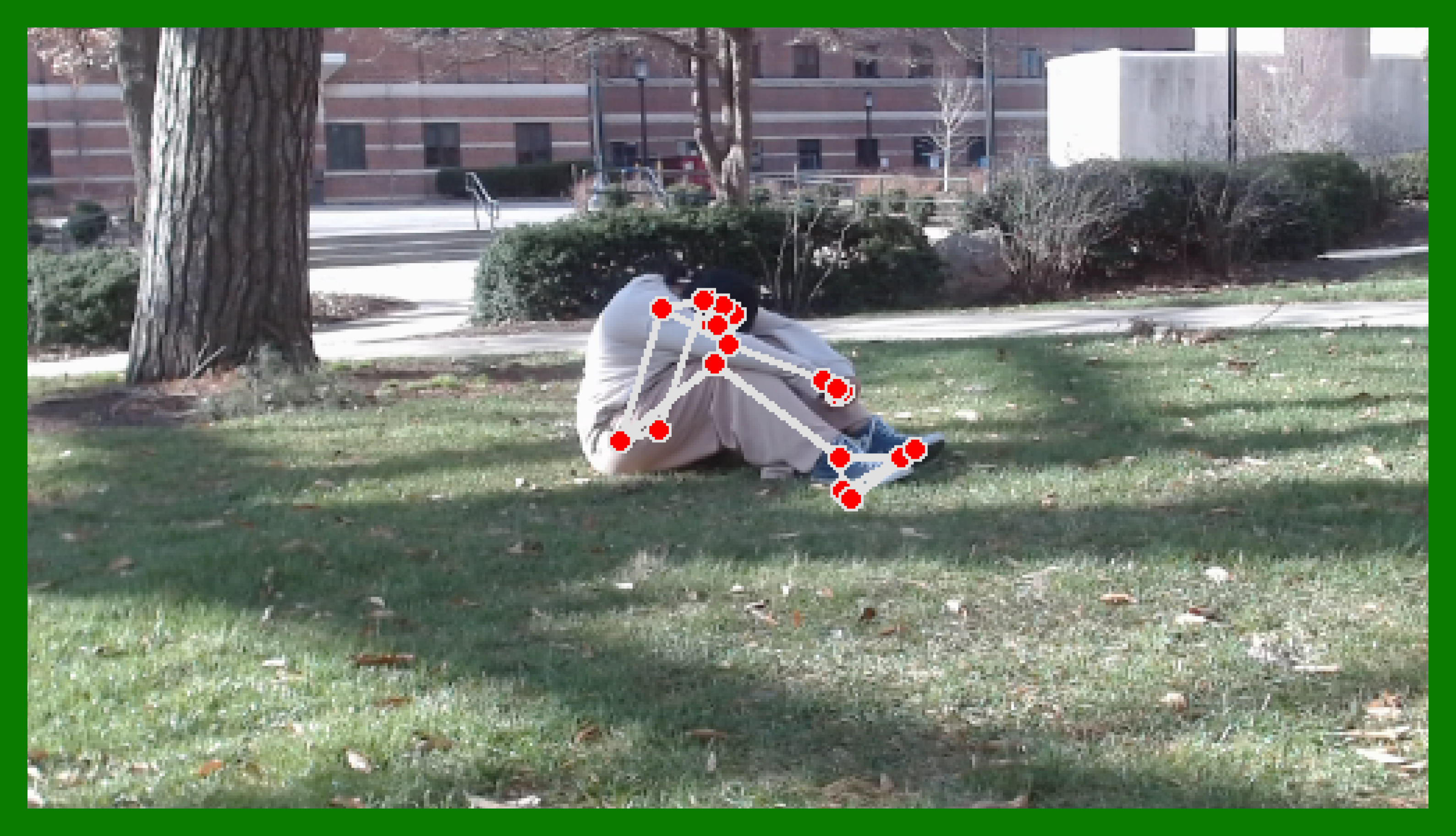}
     \caption{Far away pose}
     \label{fig:far2}
 \end{subfigure}
  \hfill
 \begin{subfigure}[b]{0.24\linewidth}
     \centering
     \includegraphics[width=\linewidth]{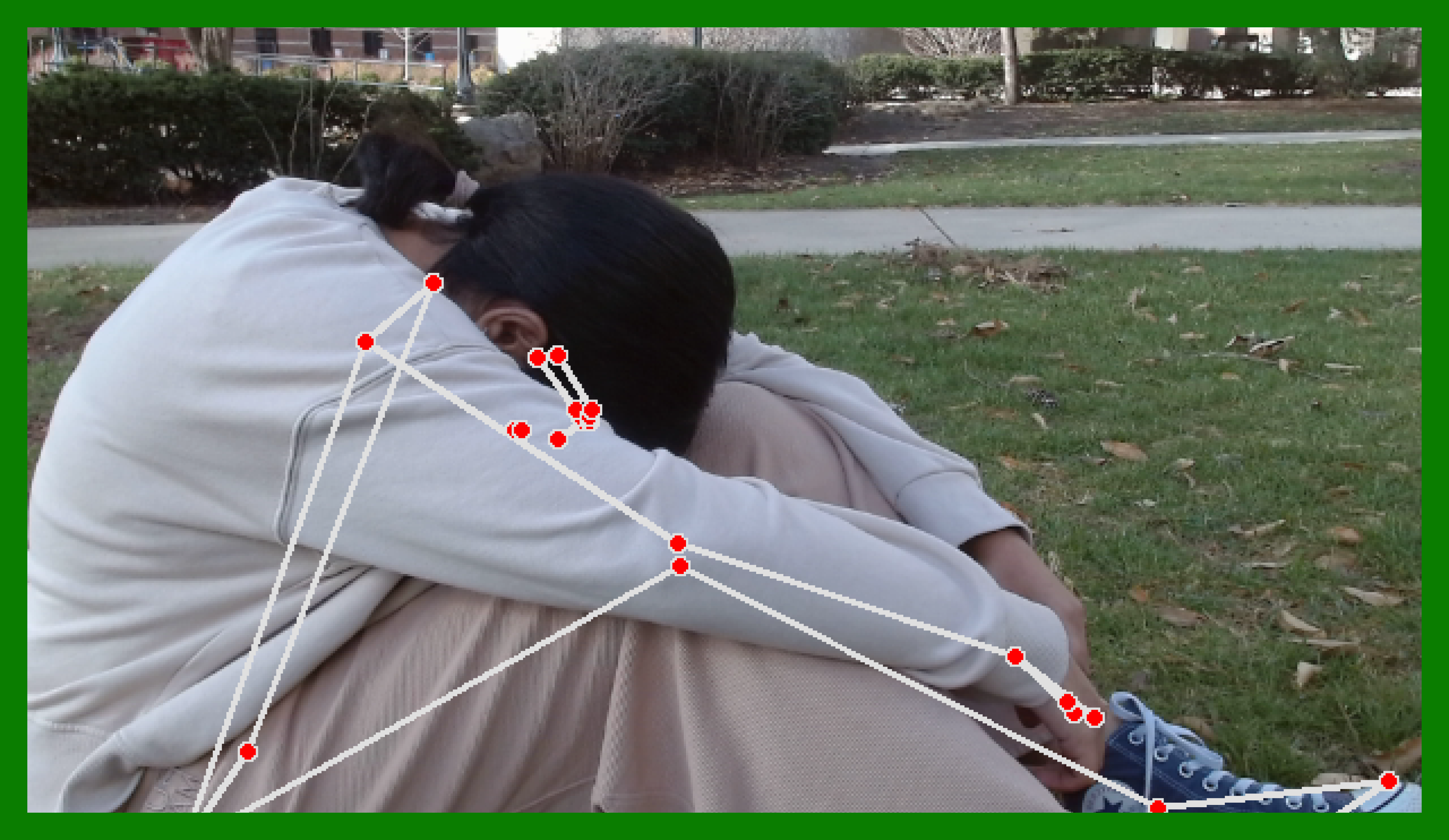}
     \caption{Close-up pose}
     \label{fig:close2}
 \end{subfigure}
\caption{\textit{An overview of the collection of vitals, triage classification, and the intervention of first responders during Mass Casualty Incidents (MCI). \textbf{ARTEMIS}, uses machine learning to rapidly and accurately identify the acuity levels of the identified victims.}}
\label{fig:protocol}
\end{figure}

During an MCI, first responders are tasked with locating all the victims and perform the preliminary assessment. Triage is defined as the process of preliminary assessment of victims in order to determine the urgency of their need for treatment and the nature of treatment required. This is a two-stage process: the primary triage, which is rapidly done on-site, and the secondary triage, which is done once the patients enter the treatment area. Leveraging robots during MCIs can help expedite the localization process since robots have better mobility due to their compact form factor \cite{adam2023using, hashimoto2017four}. Additionally, in hazardous conditions, employing robot teams for localization serves to mitigate the inherent risks faced by first responders, offering a safer and faster approach to the critical tasks at hand \cite{hoque2024hri, allison2024towards, pillai2024heterogeneous}. 

While robots have been used in emergency departments (EDs) for triage in assisting doctors and first responders to treat and get vitals from patients \cite{chai2021robot, townsend2023medical} and in MCIs for search \& rescue \cite{baxter2007multi, gonzalez2010toward}, there exists a lot of potential for developing robots that are tightly coupled with first responders in real-time. Incorporating robots into disaster response missions offers the potential to gather crucial real-time data and relay on-the-ground information. 

However, past deployment of robots in disaster settings have faced many issues, such as human operator error, imperfect autonomy, poor interface, lack of robustness, inefficient power systems, legal challenges, incorrect parameter tuning, etc. \cite{jin2019research, park2017disaster}. This work attempts to address some of these challenges. A quadruped is utilized for better mobility in rugged terrain \cite{leeRobotTerrain, fankhauser2018robust, valsecchi2020quadrupedal}. Recognizing their role as assistive tools rather than replacements for human responders in MCI scenarios, the quadruped's only function is to conduct localization and preliminary triage classification in scenarios with limited medical personnel, enabling first responders to efficiently prioritize subsequent assessments. We improve on state-of-the-art machine learning triage models to achieve better classification accuracy and provide an intuitive user interface for first responders.



Collecting vitals from humans autonomously or with the help of robots has been proven to be possible with both contact-based and contact-free sensing methods between the robot and the victim \cite{WANG20221, Huang2022SPOT, Kebe2020}. However, since the hardware necessary for such monitoring shows promise for implementation, this paper shifts its focus toward the operational strategies of robotic systems in assisting first responders in speeding up the primary triage process. Furthermore, we explore the use of machine learning and natural language processing as cost-effective approaches to complement hardware sensors for a more holistic initial triage assessment, under the assumption that in the later stages of triage, medical personnel will be more efficient in using the right diagnostic devices. We do not collect vitals from users in the proposed method, but we show that using such technologies with ARTEMIS would make this a complete system. We propose \textbf{ARTEMIS} or \textbf{A}I-driven \textbf{R}obotic \textbf{T}riage Labeling and \textbf{E}mergency \textbf{M}edical \textbf{I}nformation \textbf{S}ystem as a robot-driven primary triage framework.

To this end, we make the following contributions:

\begin{itemize}

  \item \textbf{Robot-driven Victim Identification}: The deployment of a pre-trained human detection Mediapipe BlazePose model \cite{bazarevsky2006blazepose} on an unmanned robot quadruped to identify humans in the operating environment, autonomously approach them, and send their location and trajectory to first responders. 
  
  \item \textbf{AI-based Triage Classification}: A machine learning approach to rapidly identify the triage levels of the identified victims, that is trained on a publicly available dataset collected by the  Department of Emergency Medicine, Yale School of Medicine, New Haven, Connecticut, United States of America \cite{vantu2023medical}.
   
  \item \textbf{Graphical Frontend}: A user-friendly and informative GUI for real-time communication between triage robots and first responders is created. This helps visually depict mission-critical information such as the obstacle-free path to the victim, vital signs and acuity labels, obtained from triage classification.
  
  \item \textbf{Synthetic Triage Dataset}: Y-MED-SYN+, an augmented dataset that is synthetically prepared from the Yale medical dataset \cite{vantu2023medical}, tailored for primary triage classification, is published. The dataset can be requested at \href{https://ideas.cs.purdue.edu/research/artemis/}{https://ideas.cs.purdue.edu/research/artemis/}.
\end{itemize}

This entire process is depicted in Figure \ref{fig:protocol}, which shows a quadruped scanning of the environment and approaching victims to perform initial triage classification. 



\section{RELATED WORKS}

This section highlights and analyzes some existing ways of deploying Machine Learning with triage protocols, and assesses the kinds of robotics capabilities that have come close to being used for such applications.

\subsection{Machine Learning for Mass Casualty Incident Triage}\label{subsec:global-triage-protocols}

Triage is the initial screening examination that divides the patients into multiple categories based on the severity of their injuries. This is a two-stage process, which includes primary (rapid, on-site) and secondary (treatment area) triage. However, there has been much debate in the area to identify what incidents constitute an urgency \cite{zimmermann2001case, twomey2012south, connelly2004discrete}. The ESI 5-level triage system \cite{Lisa_Katrina_Danielle_Deena_2023} assists in identifying this and is also able to describe precisely the condition of the patient\cite{travers2002five}. Machine Learning (ML) based systems have demonstrated significant potential in triaging patients more effectively in Emergency Departments (ED) by analyzing large datasets to predict the urgency of medical attention required. Studies have shown that ML algorithms can outperform traditional triage methods, facilitating rapid decision-making which is crucial in high-pressure scenarios typical of MCIs \cite{vantu2023medical, defilippo2023computational, yuba2022systematic}. 

Building on these advancements, the application of robotics equipped with ML-based triage classification in MCIs can further enhance the capabilities of first responders. Robots can navigate through unstructured environments to reach patients, assess their condition using ML algorithms, and categorize them based on the severity of their injuries or medical conditions. This automated triage system can significantly reduce the time taken for initial assessment and enable a more organized response, thereby streamlining rescue operations. By providing accurate, real-time data to first responders, these robotic systems can ensure that medical attention is directed where it's needed most, enhancing the overall efficiency of emergency response efforts \cite{yu2020machine, LEVIN2018565, Bazyar2019}.

\begin{figure*}[h!]
\centering
\includegraphics[width=1\textwidth]{pipeline.pdf}
\caption{\textit{ARTEMIS System Architecture: a Unitree Go1 quadruped to be used for collecting patient vitals (Heart Rate, Respiratory Rate, $O_2$ Sat., Systolic and Diastolic Blood Pressure, etc.), chief complaints, and age. The Multi-Layer Perceptron (MLP) uses this to classify the patient's acuity level, which is then displayed on the GUI along with the patient's location and photograph.}}
\label{fig:pipeline}
\end{figure*}

\subsection{Robot Assistance in Mass Casualty Incidents}\label{subsec:robots-in-mci}

In recent years, the use of robots for assistance in mass casualty incidents (MCIs) has garnered significant interest, demonstrating potential to revolutionize emergency response strategies. Historically, research and development in rescue robotics have laid a foundation for their application in MCIs. Notably, some works explored the concept of robot-assisted mass-casualty triage, even in marine scenarios, highlighting the potential of robots in assessing and prioritizing victims in large-scale emergency scenarios \cite{Chang4239002, Xiao8206510}. These studies provide valuable insights into the capabilities and limitations of robotics in emergency settings.

Building upon these foundational works, this paper addresses how the integration of triage classification and machine learning (ML) can enhance the effectiveness of robots in MCIs. The incorporation of ML algorithms enables robots to rapidly and accurately assess victims' medical needs, facilitating more efficient triage and prioritization. This, in turn, aids first responders in making informed decisions, ensuring that critical resources are allocated where they are most needed.  Robot-assisted emergency response education and training, which can further benefit from ML-driven triage systems, has also been in the rise recently\cite{chaudhary2023heroes}. By combining the strengths of rescue robotics with advanced ML techniques, we can envisage a future where robots not only assist in navigating and assessing MCIs but also play a crucial role in enhancing the overall response strategy, ultimately improving outcomes in these high-stakes situations.

\begin{algorithm}
\caption{Human Pose Detection and Navigation}
\begin{algorithmic}[1]
\Require Video Input Stream, $\mathbb{V}_{in}$; Pose Model, $\mathbb{M}_P$; Ultrasonic Reading, $\mathbb{U}$; Approach Threshold $\zeta_A$;
\Ensure Human Pose Landmarks, $\mathbb{L}$; Success Tag $\mbox{T}$; Robot Velocity Command, ${cmd\_vel}$
\While {$\mathbb{L} == \{\}$}\tikzmark{top}
    \State ${\mathbb{L} \gets Search(\mathbb{V}_{in}, \mathbb{M}_P)}$
\EndWhile
\While {$\mathbb{U} > \zeta_A$}
    \State $\mathbb{L} \gets \mathbb{M}_P(\mathbb{V}_{in}) $
    \State $Avg\_pos \gets Centroid(\mathbb{L})$\tikzmark{bottom}
    \State $cmd\_vel.yaw \gets Robot Yaw Correction (Avg\_pos)$\tikzmark{top1}
    \tikzmark{right}\tikzmark{right1}
    \State $cmd\_vel.linear \gets Move Forward()$
\EndWhile
\If {$\mathbb{L} != \{\}$}
    \State $T \gets SUCCESS$\tikzmark{bottom1}
\EndIf
\State {$Repeat$}

\end{algorithmic}
\label{alg:pose_alg}
\AddNote{top}{bottom}{right}{Human Pose Detection}
\AddNote{top1}{bottom1}{right1}{Navigation}
\vspace{-4mm}
\end{algorithm}

\section{Methodology}
The architecture of \textbf{ARTEMIS} includes traversing an obstacle-free path to the victim, identification of victims on site with computer vision, followed by location tagging and acuity label assignment. Thereafter, the model relays the data to the first responders through a GUI-based frontend. The following subsections provide detailed descriptions of these steps. The overall pipeline has been illustrated in Fig.~\ref{fig:pipeline}.

\subsection{Search Problem Formulation}

Victim localization is modeled as a search problem. Define an environment $\mathscr{E} \subseteq \mathbb{R}^2$ with an unknown number of targets (victims), whose locations are unknown. Let R be the robot placed in a bounded environment $\mathscr{E}$ \cite{batalin2002multi}. The process of victim localization involves traversing a Hamiltonian Path \cite{jedlivckova2023hamiltonian}. A graph G of all the N targets is defined. In this problem setting, neither the number of targets (N) or their locations are known. The robot executes a Hamiltonian walk h(G) on graph G, visiting each target once. Given that the graph may have multiple edges indicating alternate paths between the pair of targets, the problem of finding a Hamiltonian path is NP-complete \cite{jedlivckova2023hamiltonian}. When the global information of the graph, i.e. locations of all the N targets and all possible paths are known, this becomes the orienteering problem \cite{blum2007approximation}, where the minimum cost of reaching every target, as a function of the distance between two targets, is denoted by
\begin{equation}
    min \sum_{i=1}^{N} \ \sum_{j \neq i, j=1}^{N} c_{ij} x_{ij}
\label{eq:cost}
\end{equation}
In equation \ref{eq:cost}, $c_{ij}$ is the cost of traversing between the two targets, which could be a function of distance or battery usage of the robot, and $x_{ij}$ is an indicator variable denoting whether an edge exists between i and j. Using dynamic programming, the problem can be solved with a time complexity of $\mathbf{O}({N}^2 2^{N})$.

\subsection{Navigation and Trajectory Control}
 When the number of targets is unknown, at the start of the search, the robot R executes a heuristic search  \cite{heurReview}. This can be defined with farthest point sampling, where we have a set of explored points ($\mathscr{S}$), which is a subset of all the points in the environment ($\mathscr{E}$). In case of localization, these points include the victims that the robot has attended to. The algorithm maximizes the distance between the set of explored points and a randomly chosen point from the environment. The algorithm can be represented as 
\begin{equation}
    x_{fps} = \argmax_{x\in\mathscr{E}, \ s\in\mathscr{S}} ||x - s||^2_2
\end{equation}
 Due to a lack of information about N, the algorithm's termination is determined by the area covered based on the tracked trajectory, where robot R maintains a hierarchical list of points explored and stops the search when the entire area is covered \cite{JOHO2011319}.
 
\subsection{Patient detection and localization}

As the initial step, the quadruped employs visual perception through a camera (connected to an external computer) mounted on it, to conduct a scan of the environment by searching for humans. The captured frames are transmitted to a pre-trained human pose detection model - Media Pipe BlazePose \cite{bazarevsky2006blazepose} implemented on the external computer, where real-time analysis takes place. This model detects and annotates the pose and presence of humans within the observed scene. Subsequently, based on the position of the detection, the robot corrects its heading and approaches the subject to then proceed with acquiring their vitals. Algorithm \ref{alg:pose_alg} summarizes the process of detection and approach. 

\subsection{Location tagging the victim}

Apart from the acuity label assignment, it is important to share the victim's location with the emergency responders to make it easier to reach the victims. It is also important to share the path the robot takes to reach the victim because this path can be assumed to be free from obstacles such as rubble. This is done with the help of an onboard inertial measurement unit (IMU) that helps update the robot's trajectory as it moves. 
\subsection{Triage Classification}

\label{subsection B}

Following the detection of victims, the next step is the preliminary assessment by the quadruped. The quadruped initiates a series of questions asking the victim to state their primary symptoms. Victims undergo the primary triage process, wherein the assigned label is determined partly by their responses (chief complaints such as ``knee pain", ``chest pain", or ``numbness"). The acuity labels range from 1 to 5, with acuity label 1 indicating the most critical condition, such as low vitals or being unresponsive and acuity label 5 indicates that the victim is fairly stable. Additionally, the quadruped may be equipped with non-invasive contact sensors capable of gathering vital signs, including pulse, oxygen saturation, temperature and blood pressure.
\subsection{Human-Robot-Interaction}

To effectively interact with humans, Python's \texttt{Speech Recognition} and \texttt{pyttsx3} packages, with which it can communicate verbally, are deployed on the robot. A standard American English female voice is used since the quadruped was tested in the United States. However, it is possible to change the accent to the region in which the robot is deployed to ensure that victims have no difficulty understanding the robot. At the moment, ARTEMIS only supports the English language. 

\subsection{Graphical User Interface (GUI)}

A key feature for communicating with first responders is a graphical user interface (GUI) that updates victim information in real-time. The victim's chief complaints, acuity label, trajectory, and location are transmitted from the robots on-site to a central server, which can be accessed from a web-based front end.

\section{Experiments and Results}


\begin{table}[ht]
\resizebox{0.48\textwidth}{!}{%
\begin{tabular}{|c|c|c|c|c|c|c|c|c|}
\hline
\textbf{Dataset}  & \textbf{Model} & \textbf{Overall Accuracy} & \textbf{Class} & \textbf{Precision} & \textbf{Recall} & \textbf{F1 Score} & \textbf{Accuracy} \\ \hline
\multirow{5}{*}{MIMIC}  & \multirow{5}{*}{\begin{tabular}[c]{@{}c@{}}5-Layer\\ MLP (Proposed Model)\end{tabular}} & \multirow{5}{*}{59} 
    & \cellcolor[HTML]{EFEFEF}1 & \cellcolor[HTML]{EFEFEF}0.80 & \cellcolor[HTML]{EFEFEF}0.64 & \cellcolor[HTML]{EFEFEF}0.71 & \cellcolor[HTML]{EFEFEF}0.64 \\ 
   &  &  & 2 & 0.43 & 0.35 & 0.39 & 0.35 \\ 
   &  &  & \cellcolor[HTML]{EFEFEF}3 & \cellcolor[HTML]{EFEFEF}0.34 & \cellcolor[HTML]{EFEFEF}0.41 & \cellcolor[HTML]{EFEFEF}0.37 & \cellcolor[HTML]{EFEFEF}0.41 \\ 
   &  &  & 4 & 0.52 & 0.57 & 0.54 & 0.57 \\  
   &  &  & \cellcolor[HTML]{EFEFEF}5 & \cellcolor[HTML]{EFEFEF}0.90 & \cellcolor[HTML]{EFEFEF}\textbf{0.98} & \cellcolor[HTML]{EFEFEF}\textbf{0.94} & \cellcolor[HTML]{EFEFEF}\textbf{0.98} \\ \hline
\multirow{5}{*}{\begin{tabular}[c]{@{}c@{}}\textbf{Yale}\\\textbf{Dataset (Y-MED)}\end{tabular}}  & \multirow{5}{*}{\begin{tabular}[c]{@{}c@{}}\textbf{5-Layer}\\ \textbf{MLP (Proposed Model)}\end{tabular}} & \multirow{5}{*}{\textbf{74}} 
    & 1 & \textbf{0.98} & 0.99 & \textbf{0.99} & 0.99 \\  
   &  &  & \cellcolor[HTML]{EFEFEF}2 & \cellcolor[HTML]{EFEFEF}0.76 & \cellcolor[HTML]{EFEFEF}\textbf{0.71} & \cellcolor[HTML]{EFEFEF}0.74 & \cellcolor[HTML]{EFEFEF}\textbf{0.71} \\  
   &  &  & 3 & 0.62 & \textbf{0.55} & 0.58 & \textbf{0.55} \\  
   &  &  & \cellcolor[HTML]{EFEFEF}4 & \cellcolor[HTML]{EFEFEF}0.61 & \cellcolor[HTML]{EFEFEF}\textbf{0.69} & \cellcolor[HTML]{EFEFEF}0.65 & \cellcolor[HTML]{EFEFEF}\textbf{0.69} \\  
   &  &  & 5 & 0.74 & 0.75 & 0.74 & 0.75 \\ \hline
\multirow{5}{*}{\begin{tabular}[c]{@{}c@{}}Yale\\Dataset (Y-MED)\end{tabular}} & \multirow{5}{*}{\begin{tabular}[c]{@{}c@{}}Random\\ Forest\end{tabular}} & \multirow{5}{*}{73} 
    & \cellcolor[HTML]{EFEFEF}1 & \cellcolor[HTML]{EFEFEF}0.53 & \cellcolor[HTML]{EFEFEF}0.99 & \cellcolor[HTML]{EFEFEF}0.69 & \cellcolor[HTML]{EFEFEF}0.99 \\  
   &  &  & 2 & \textbf{0.87} & 0.68 & \textbf{0.76} & 0.68 \\  
   &  &  & \cellcolor[HTML]{EFEFEF}3 & \cellcolor[HTML]{EFEFEF}\textbf{0.78} & \cellcolor[HTML]{EFEFEF}0.50 & \cellcolor[HTML]{EFEFEF}\textbf{0.61} & \cellcolor[HTML]{EFEFEF}0.50 \\  
  &  &  & 4 & \textbf{0.81} & 0.58 & \textbf{0.68} & 0.58 \\  
   &  &  & \cellcolor[HTML]{EFEFEF}5 & \cellcolor[HTML]{EFEFEF}\textbf{0.91} & \cellcolor[HTML]{EFEFEF}0.88 & \cellcolor[HTML]{EFEFEF}0.90 & \cellcolor[HTML]{EFEFEF}0.88 \\ \hline
\multirow{5}{*}{\begin{tabular}[c]{@{}c@{}}Yale\\Dataset (Y-MED)\end{tabular}} & \multirow{5}{*}{\begin{tabular}[c]{@{}c@{}}Gaussian\\ Naive Bayes\end{tabular}} & \multirow{5}{*}{40} 
    & 1 & 0.35 & \textbf{1} & 0.51 & \textbf{1} \\  
   &  &  & \cellcolor[HTML]{EFEFEF}2 & \cellcolor[HTML]{EFEFEF}0.56 & \cellcolor[HTML]{EFEFEF}0.12 & \cellcolor[HTML]{EFEFEF}0.20 & \cellcolor[HTML]{EFEFEF}0.12 \\  
   &  &  & 3 & 0.58 & 0.07 & 0.12 & 0.07 \\  
   &  &  & \cellcolor[HTML]{EFEFEF}4 & \cellcolor[HTML]{EFEFEF}0.50 & \cellcolor[HTML]{EFEFEF}0.12 & \cellcolor[HTML]{EFEFEF}0.19 & \cellcolor[HTML]{EFEFEF}0.12 \\  
   &  &  & 5 & 0.46 & 0.70 & 0.56 & 0.70 \\ \hline
\multirow{5}{*}{\begin{tabular}[c]{@{}c@{}}Yale\\Dataset (Y-MED)\end{tabular}} & \multirow{5}{*}{\begin{tabular}[c]{@{}c@{}}SVM\end{tabular}} & \multirow{5}{*}{38} 
    & \cellcolor[HTML]{EFEFEF}1 & \cellcolor[HTML]{EFEFEF}0.48 & \cellcolor[HTML]{EFEFEF}0.61 & \cellcolor[HTML]{EFEFEF}0.54 & \cellcolor[HTML]{EFEFEF}0.61 \\  
   &  &  & 2 & 0.39 & 0.33 & 0.36 & 0.33 \\  
   &  &  & \cellcolor[HTML]{EFEFEF}3 & \cellcolor[HTML]{EFEFEF}0.32 & \cellcolor[HTML]{EFEFEF}0.09 & \cellcolor[HTML]{EFEFEF}0.14 & \cellcolor[HTML]{EFEFEF}0.09 \\  
   &  &  & 4 & 0.29 & 0.23 & 0.26 & 0.23 \\  
   &  &  & \cellcolor[HTML]{EFEFEF}5 & \cellcolor[HTML]{EFEFEF}0.35 & \cellcolor[HTML]{EFEFEF}0.63 & \cellcolor[HTML]{EFEFEF}0.45 & \cellcolor[HTML]{EFEFEF}0.63 \\ \hline
\end{tabular}%
}
\caption{The table shows the statistical analysis of different Machine Learning techniques applied on two datasets - MIMIC and Y-MED. It is to be noted that both the datasets are standardized and augmented using SMOTE. The MLP, with an overall accuracy of 74\%, outperforms other models trained.}
\label{tab:table1}
\end{table}

In this section, the datasets used to train the triage classifier are described (Sec.~\ref{subsec:dataset}), followed by the data pre-processing steps taken (Sec.~\ref{subsec:pre-processing}). The ML models evaluated are described and compared against state-of-the-art models in Section \ref{subsec:modelz}. The specifications of the robot prototype used for field trials (Sec.~\ref{subsec:robot-trials}) are explained (Sec.~\ref{subsec:dataset}), followed by the performance evaluation (Sec.~\ref{subsec:evaluation}). 

\subsection{Dataset}\label{subsec:dataset}
To train the machine learning models, a de-identified dataset collected by the Yale School of Medicine \cite{vantu2023medical} is used. This is referred to as the ``Y-MED'' (Yale Medical Dataset). This is a publicly accessible dataset containing patient medical data, which was collected from the emergency department in the Yale New Haven health system between March 2014 and July 2017. The Y-MED contains patient age, vital measurements (temperature, heart rate, $O_2$ saturation, age), chief complaints (knee pain, chest pain etc), and medical history in addition to the acuity level, which follows the ESI scale. In order to predict the acuity level of patients in a mass casualty incident, vitals, age, and chief complaints are the only attributes of relevance. The MIMIC-IV-ED triage dataset from BIDMC \cite{johnson2023mimic} was also used to train models. This dataset (referred to here as ``MIMIC'') only contains data for the six vital signs.

\subsection{Data Pre-processing}\label{subsec:pre-processing}

Out of the original 972 attributes in Y-MED, only 207 of them are deemed relevant. After removing outliers and empty records based on \cite{Lisa_Katrina_Danielle_Deena_2023}, the dataset is left with 268,469 of the original 560,486 records. Of the original 425,087 records in MIMIC, 224,736 records remained after applying the same cleaning process.

To help the models learn better and to improve the generalizability of the model, the records are re-scaled using z-score normalization, using the mean and standard deviation of the cleaned dataset.

Additionally, both the cleaned MIMIC and Y-MED have a significant class imbalance. In MIMIC, only 0.21\% of the dataset represent patients with acuity level 5 and 53\% represent patients with acuity level 3. In the Y-MED, only 0.1\% represent patients with acuity level 1 and 44\% represent patients with acuity level 3. To equalize the distribution, data is generated synthetically using Synthetic Minority Over-sampling Technique (SMOTE) \cite{chawla2002smote}. Each of the undersampled classes is augmented with synthetic samples generated along the line segments joining the k-nearest neighbors of samples within the class. After oversampling, Y-MED has 589,260 records and MIMIC has 593,170. The augmented and class-balanced Y-MED dataset is published as the Y-MED-SYN+ dataset.

\subsection{Models}\label{subsec:modelz}
In the following subsections, the machine learning architectures that were used for triage classification are discussed. The input for each model is a 207 attribute vector acquired from the victim by the robot. 

\subsubsection{Multi-Layer Perceptron (Proposed)}
The multi-layer perceptron (MLP) is built using TensorFlow. The network has four hidden layers, each with 50 units and a bias unit. Each of these layers uses the Rectified Linear Unit (ReLU) activation function. The output layer has five units, as well as the Softmax activation. We use Adamax Optimization with a learning rate of 0.01 and a weight decay of 1e-6. Categorical cross entropy is used as the loss function. The MLP is trained for 5000 steps. The weights of the model at the highest validation accuracy are saved and used for prediction. 

\subsubsection{Random Forest}
Python's \texttt{scikit-learn} implementation of random forest with 100 estimators is used. Bootstrapping is also employed within the forest to improve the classification accuracy. The Gini Index is used as the criteria for feature splitting.

\subsubsection{Gaussian Naive Bayes}
The Gaussian Naive Bayes approach is considered due to the presence of many continuous variables such as $O_2$ saturation, temperature etc. This algorithm assumes the attributes follow the Gaussian distribution and computes the probability as 
\begin{equation}
    P(X|Y=c) = \frac{1}{\sqrt{2\pi\sigma_c^2}}e^{\frac{-(x-\mu_c)}{2\sigma_c^2}}
\end{equation}
Where, $\mu_c$ and $\sigma_c$ are the mean and variance of a class c. The model is implemented with Python's \texttt{scikit-learn} package. 

\subsubsection{Support Vector Machine}
For multi-class classification, a one vs. one support vector machine model with the radial basis function kernel scaled by $\frac{1}{M * \sigma_c}$ is built, where M is the number of features and $\sigma_c$ is the variance of the feature. A regularization of 1 is used. This is created with Python's \texttt{scikit-learn} package.

\subsection{Go1 Quadruped Robot}\label{subsec:robot-trials}

\begin{figure}[h!]
\centering
\hspace{-4mm}
\includegraphics[width=0.5\textwidth]{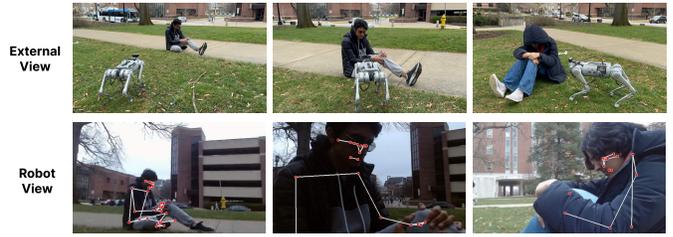}
\caption{\textit{Example of robot demo run with an external view of the robot moving (top) and a camera view from the robot's perspective (bottom). The captures from what the robot is seeing are sent over to first responders via the GUI}.}
\label{fig:final_robot}
\end{figure}

\begin{figure}
     \centering
     \begin{subfigure}[b]{0.49\linewidth}
         \centering
         \caption{SVM}
         \includegraphics[width=\linewidth]{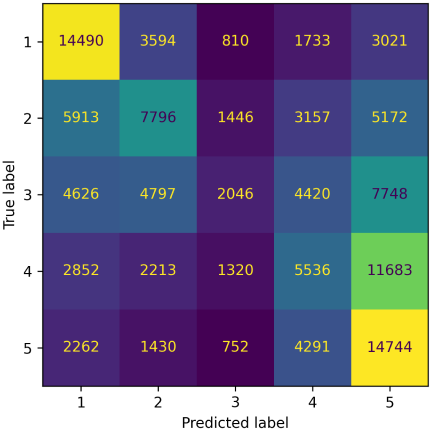}
         \label{fig:cSVM}
     \end{subfigure}
     \hfill
     \begin{subfigure}[b]{0.49\linewidth}
         \centering
         \caption{Gaussian Naive Bayes}
         \includegraphics[width=\linewidth]{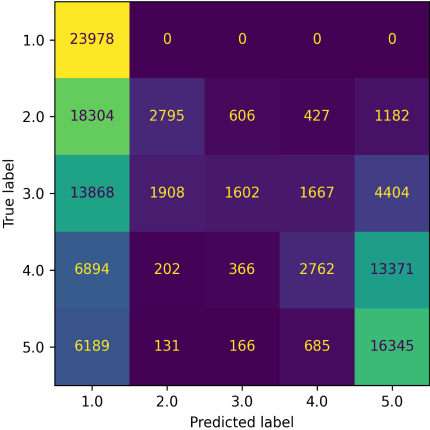}
         \label{fig:cNB}
     \end{subfigure}

    \vspace{-10pt}
     
     \hfill
     \begin{subfigure}[b]{0.49\linewidth}
         \centering
         \caption{Random Forest}
         \includegraphics[width=\linewidth]{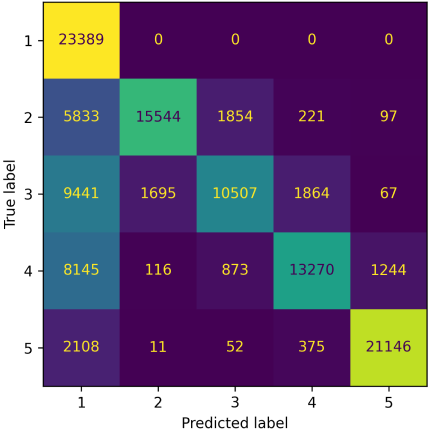}
         \label{fig:cRF}
     \end{subfigure}
     \hfill
     \begin{subfigure}[b]{0.49\linewidth}
         \centering
         \caption{MLP}
         \includegraphics[width=\linewidth]{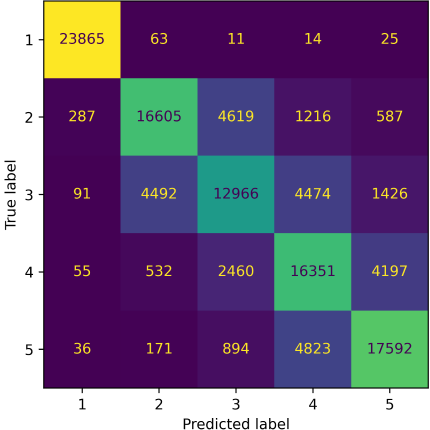}
         \label{fig:cMLP}
     \end{subfigure}
    \caption{This figure highlights the tendency of the models to incorrectly classify a triage level. Most of the incorrect labels for d) MLP are adjacent to the true label (83.4\% of all mis-classified labels).}
    \label{fig:grid}
\end{figure}

Fig.~\ref{fig:pipeline} shows the overall system architecture. The quadruped is a Unitree Go1 with an external Logitech C920 RGB camera mounted onto it. The robot control is set up in Robot Operating System 1 (ROS1) and communicates with the robot's onboard Raspberry Pi 4. A Mediapipe Blazepose pose detection algorithm helps the robot detect human pose in the environment. As the robot moves and searches for a human pose, it uses a proportional gain based yaw correction algorithm to adjust its position and approach humans in the environment as detailed in Algorithm \ref{alg:pose_alg}. The robot is also equipped with ultrasonic range sensors that make sure the robot avoids obstacles on the way and is able to approach humans and stop within a threshold distance. The robot's IMU keeps track of the trajectory of the robot, and these are shared with first responders via a Graphical User Interface (GUI).

A controlled, simulated demonstration of the victim detection, localization, and triage classification is conducted in an outdoor environment with the quadruped, as shown in Figure \ref{fig:final_robot}.

\begin{figure}[htb]
\centering
\includegraphics[width=8cm]{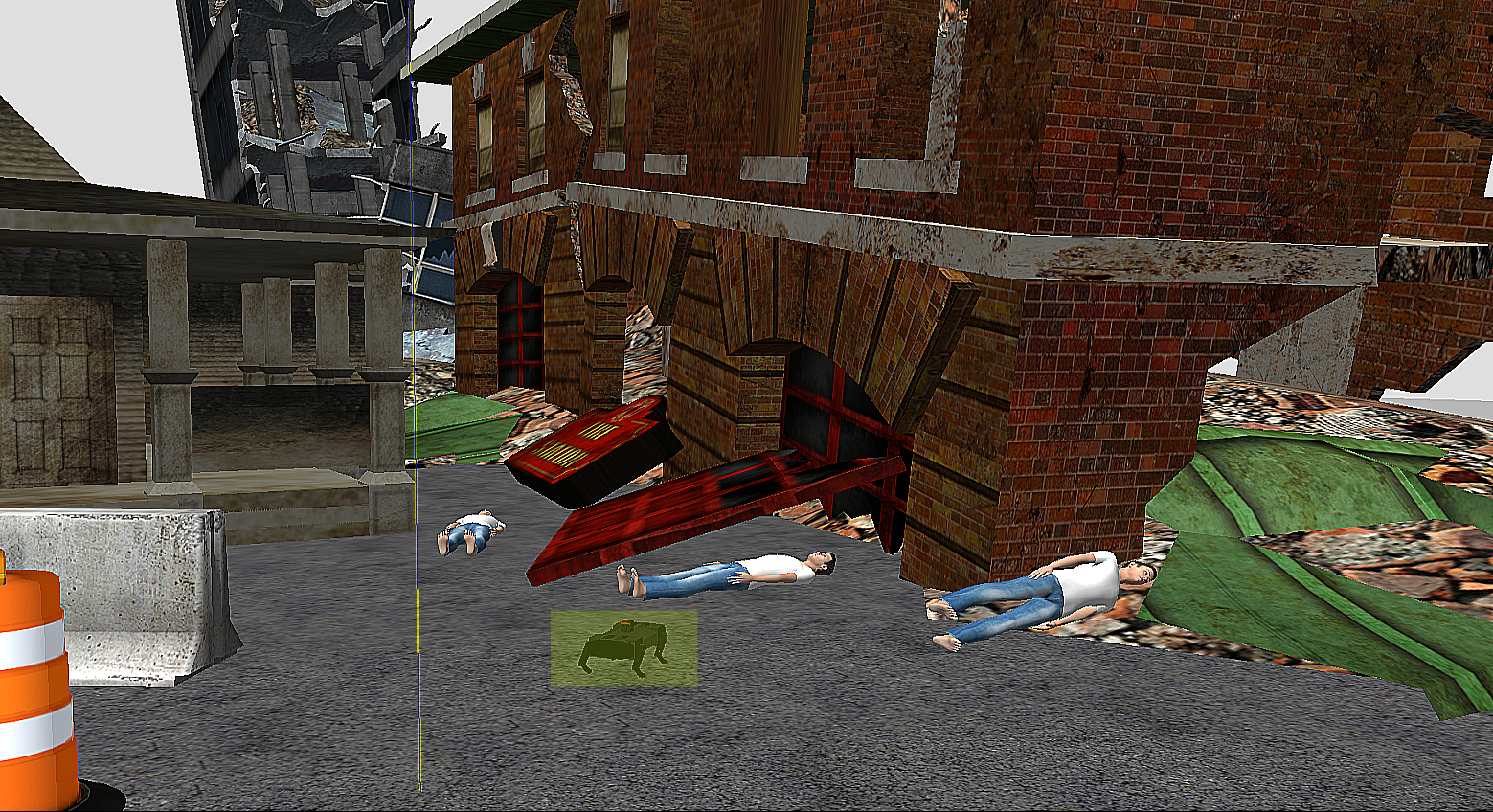}
\caption{\textit{A simulated MCI environment in Gazebo depicting victim detection is shown. This is done using CHAMP quadruped framework in ROS1 \cite{lee2013hierarchical}.}}
\label{fig:gazebo}
\end{figure}

\begin{table}[ht]

\begin{tabular}{|c|c|c|c|c|}
\hline
\textbf{Model}  & \textbf{Technique} & \textbf{Precision} & \textbf{Recall} & \textbf{F1 score}\\ \hline
\textbf{ARTEMIS}  & \begin{tabular}[c]{@{}c@{}}\textbf{5-layer MLP}\\ \textbf{(Proposed)}\end{tabular} & \textbf{0.74} & \textbf{0.74} & \textbf{0.74}\\
eUPU \cite{vantu2023medical} & Neural network & 0.71 & 0.69 & 0.70\\
eUPU \cite{vantu2023medical} & Log. Regression & 0.65 & 0.41 & 0.50\\
eUPU \cite{vantu2023medical} & Random Forest & 0.68 & 0.64 & 0.66\\
KTAS \cite{choi2019machine} & Log. Regression & 0.71 & 0.71 & 0.71\\
KTAS \cite{choi2019machine} & Random Forest & 0.73 & 0.73 & 0.73\\
KATE \cite{ivanov2021improving} & XGBoost & 0.72 & 0.67 & 0.69\\
ED-Adm. &&&&\\(Triage) \cite{hong2018predicting} & XGBoost & 0.66 & 0.69 & 0.67\\
ED-Adm. &&&&\\(Triage) \cite{hong2018predicting} & Log. Regression & 0.65 & 0.68 & 0.66\\
ED-Adm. &&&&\\(Triage) \cite{hong2018predicting} & Log. Regression & 0.66 & 0.70 & 0.68\\
\hline

\end{tabular}
\caption{Results highlighting that ARTEMIS outperforms state-of-the-art models in acuity labeling.}
\label{tab:table2}
\end{table}

\subsection{Evaluation}\label{subsec:evaluation}


Precision, recall, F1 score, and accuracy are used to evaluate the models, as shown in table \ref{tab:table1}. The MLP model, trained on Y-MED dataset, has the highest precision as well as accuracy in classifying acuity level 1, which represents the most critical patients. However, due to class imbalance, it performs poorly on MIMIC dataset in classifying all classes except acuity level 5. Gaussian Naive Bayes has 100\% accuracy in classifying acuity level 1 patients (see figure \ref{fig:cNB}) but has a very low precision for the same class. Figure \ref{fig:cMLP} shows the mis-classification tendencies of the MLP model. In Figure \ref{fig:cMLP}, it can be observed that the majority of the mis-classified victims were classified into a class that was adjacent to the true class.

Random forest has the best precision in identifying patients with acuity levels 2, 3, 4, and 5, as seen from figure \ref{fig:cRF}. However, the random forest has a lower recall than the MLP, indicating that there are a lot of false negatives. As shown in Figure \ref{fig:cSVM}, muli-class SVM has the overall lowest performance among all models.  


ARTEMIS hence deploys a 5-layer MLP with the augmented Y-MED dataset, which also has better results than the baseline models compared against. Table \ref{tab:table2} shows the performance of ARTEMIS against other state-of-the-art machine learning-based triage classification models.

\subsection{Discussion}
\label{sec:limi}


ARTEMIS shows how robots can be used to assist first responders in the process of attending to victims in MCIs. The use of sensors for vital signs acquisition can help improve the accuracy of triage classification; however, it is to be noted that these sensors will have additional battery overhead \cite{tiwari2019unified}. With the pre-processing done on the dataset used, class 1, which is the most severe acuity class, may not be representative of the ideal case as the SMOTE up-sampling increases the number of data points in class 1 significantly. Additionally, deploying the quadruped in real MCIs is not possible due to research constraints and therefore, as shown in Figure \ref{fig:gazebo}, localization within a simulated MCI environment in Gazebo was performed. The environment is set up using CHAMP framework in ROS1 \cite{lee2013hierarchical}.

\section{Conclusion}
\label{sec:conclusion}
ARTEMIS, a framework for MCIs to automate primary triage classification with mobile robots is presented. A quadruped equipped with pose detection models that can perform preliminary triage classification is shown, outperforming state-of-the-art machine learning based triage classification models. ARTEMIS outperforms state-of-the-art acuity labeling models and shows promise for integrating robots to assist in the work of first responders in an MCI. 

With future work, while current research employs a single quadruped disaster response setting, it necessitates the adoption of a heterogeneous system comprising both Unmanned Ground Vehicles (UGVs) and Unmanned Aerial Vehicles (UAVs) for improved efficiency. The primary challenge in such a system lies in devising effective power management strategies to optimize performance. Incorporating models such as DREAM \cite{patel2023dream}, which discusses the development of efficient power management strategies in heterogeneous setups, would be a preliminary step. A potential avenue for future exploration involves integrating heterogeneous coordination strategies, particularly leveraging machine learning for target localization in previously unseen environments, as proposed in \cite{peng2023graph, pham2023crowd}. The incorporation of such methodologies allows ARTEMIS to be transformed into a decentralized and more adaptive robotic framework. 

\vspace{-2mm}
\bibliographystyle{IEEEtran}
\bibliography{root}
\end{document}